\documentclass[sigconf]{acmart}
\usepackage{multirow}
\usepackage{stfloats}
\usepackage{balance}
\makeatletter
\let\@authorsaddresses\@empty
\renewcommand\@formatdoi[1]{\ignorespaces}
\makeatother
\AtBeginDocument{%
  \providecommand\BibTeX{{%
    \normalfont B\kern-0.5em{\scshape i\kern-0.25em b}\kern-0.8em\TeX}}}

\copyrightyear{2020}
\acmYear{2020}
\setcopyright{acmcopyright}
\acmConference[ACSW 2020]{Proceedings of the Australasian Computer Science Week Multiconference}{February 4--6, 2020}{Melbourne, VIC, Australia}
\acmBooktitle{Proceedings of the Australasian Computer Science Week Multiconference (ACSW 2020), February 4--6, 2020, Melbourne, VIC, Australia}
\acmPrice{15.00}
\acmDOI{10.1145/3373017.3373049}
\acmISBN{978-1-4503-7697-6/20/02}

\begin{document}

\title{HHH: An Online Medical Chatbot System based on Knowledge Graph and Hierarchical Bi-Directional Attention}


\author{Qiming Bao}
\authornote{Corresponding author}
\email{qbao775@aucklanduni.ac.nz}
\orcid{https://orcid.org/0000-0002-1000-7383}
\affiliation{%
  \institution{The University of Auckland}
  \streetaddress{Building 303 & 303S, 38 Princes Street, City Campus}
  \city{Auckland}
  \state{Auckland}
  \postcode{1010}
}

\author{Lin Ni}
\email{l.ni@auckland.ac.nz}
\orcid{https://orcid.org/0000-0001-7421-0623}
\affiliation{%
  \institution{The National Institute for Health Innovation(NIHI)}
  \streetaddress{Level 1, Building 507, Grafton Campus85 Park Road, Grafton}
  \city{Auckland}
  \state{Auckland}
  \postcode{1023}
}

\author{Jiamou Liu}
\email{jiamou.liu@auckland.ac.nz}
\orcid{https://orcid.org/0000-0002-0824-0899}
\affiliation{%
  \institution{The University of Auckland}
  \streetaddress{Building 303 \& 303S, 38 Princes Street, City Campus}
  \city{Auckland}
  \state{Auckland}
  \postcode{1010}
}

\renewcommand{\shortauthors}{Qiming Bao, Lin Ni and Jiamou Liu}

\begin{abstract}
This paper proposes a chatbot framework that adopts a hybrid model which consists of a knowledge graph and a text similarity model. Based on this chatbot framework, we build HHH, an online question-and-answer (QA) Healthcare Helper system  for answering complex medical questions. 
HHH maintains a knowledge graph constructed from medical data collected from the Internet. HHH also implements a novel text representation and similarity deep learning model, Hierarchical BiLSTM Attention Model (HBAM), to find the most similar question from a large QA dataset. We compare HBAM with other state-of-the-art language models such as bidirectional encoder representation from transformers (BERT) and Manhattan LSTM Model (MaLSTM). We train and test the models with a subset of the Quora duplicate questions dataset in the medical area. 
The experimental results show that our model is able to achieve a superior performance than these existing methods. 
\end{abstract}




\maketitle
\thispagestyle{empty}

\section{Introduction}

Difficulty in seeing a doctor, long queuing time, and inconvenience of making appointments have long been hurdles facing patients when they try to access primary care services. To solve these challenges, governments and health care providers around the world are investing in new methods that facilitate more effective use of resources to meet demands. As an example, New Zealand government has issued  the ``6-hour target'' in 2009 aiming to significantly boost the availability of medical resources \cite{jones2012implementing}, while more recently, the Precision Driven Health initiative targets a new model that joins force government, commercial, health care providers and researchers in New Zealand, in the hope to better harness the power of digital medical data and information technology to deliver enhanced services \cite{dobbie2017precision}. 

Artificial intelligence plays a crucial role in the advancement of information technology to improve healthcare service quality and efficiency. 
In particular, chatbots amount to one of the most popular AI technologies for this purpose. A chatbot is a software system that consists of an interactive interface with patients or medical practitioners to provide a range of knowledge extraction tasks and real-time, personalized feedback. Chatbot technologies have been rapidly developed, especially in the medical field. Many medical chatbot systems have been proposed over the years. Typical applications of chatbot include medical assistants that help patients to identify their symptoms, medical service front desks that direct the patient to suitable healthcare service departments, i.e., doctors, and so on.

Our work aligns with the main themes of medical chatbot technology and aims to serve three main objectives: The first objective is to reduce waste on resources and time for users when accessing information with chatbot technologies. We aim to maximally help users to search for the necessary information with a human-like interface. The second objective is to provide more precise answers to ordinary users who have little domain knowledge. In other words, we hope that with AI technologies, the system can understand the meaning of the natural language and be able to reply with high-quality feedback accordingly. The third objective is to make it easier to manage and extend the features and databases. We want to design a system with a flexible and scalable structure to enable efficient management of the functionality and datasets. 

To this end, we first design a framework to implement a generic chatbot system.  Our chatbot framework contains two main modules. The first module is the user interface, which contains a web-based chatbot front-end, a local GUI, and a back-end to handle database management. The second module serves to respond to user's queries based on our hybrid QA model, which contains a knowledge graph and the {\em hierarchical BiLSTM attention model} (HBAM). 

We build our {\em Healthcare Helper system with a Hybrid QA model} (HHH) as an instance of the chatbot framework above. The knowledge graph stores more than 600 different kinds of disease records and is able to answer six different types of questions, while the HBAM can query from a big dataset containing 29287 medical questions-and-answer pairs (171 from ehealthforumQAs, 5679 from questionDoctorQAs and 23437 from webmdQAs).

One novelty of our work lies in the utilization of a {\emph{hybrid QA model}} that combines a knowledge graph database and an NLP model. A user's question firstly will be queried from the knowledge graph. If it cannot find any result, a text similarity model will be used to find the answers from a large medical QA dataset. 

The highlight of this paper within this model involves a novel deep learning-based text-representation and similarity-comparison model: the {\emph{HBAM}}. HBAM consists of a BiLSTM layer and a word attention layer. The functionality of the BiLSTM layer is to capture the forward and reverse directional information of a sentence. The word attention layer is used to capture the keywords in a sentence. Siamese framework and Manhattan distance are used to compute the medical level semantic similarity. Siamese framework has been widely proposed in the metric learning tasks \cite{yih2011learning} \cite{chen2011extracting}. Manhattan distance has been utilized to measure sentence similarity, such as cosine similarity \cite{yih2011learning}. Comparing with MaLSTM \cite{mueller2016siamese} and BERT \cite{devlin2018bert}, our HBAM gets the highest score in the experiments with different datasets.

\paragraph*{\bf \bf Paper organization.} The rest of the paper is organized as follows. Section~\ref{sec:problem} presents the two core problems studied in this paper and presents related works. Section~\ref{sec:system} presents the main system architecture of the chatbot framework. Section~\ref{sec:KG} presents how the knowledge graph is implemented for our medical chatbot. Section~\ref{sec:HBAM} describes our HBAM model which is the key to natural language understanding. Section~\ref{sec:results} provides some sample output of the system as well as a quantitative analysis of the performance of the system using two sets of experiments. The results show that our system achieves superior performance as compared to existing systems. Section~\ref{sec:conclusion} concludes the paper with a discussion of potential future work. 

\section{Problem Formulation and Related Work}\label{sec:problem}

In the following, we define two problems that are at the center of the chatbot system. The first problem aims to realize the ability of natural language understanding, i.e., developing the necessary mechanism for the software system to understand natural language questions as a human would do. The second problem aims to extract the relevant information from a domain-specific database so that answers can be generated to be fed back to the user. 
\begin{enumerate}
\item User question understanding (Intent Detection): Natural language understanding (NLU) and natural language processing (NLP) to understand and process a user's question. 
\item Knowledge base storage and retrieval: A domain knowledge database to be able to store and query the medical questions and answers.
\end{enumerate}

\begin{figure*}[t]
\centering
\includegraphics[width=0.8\textwidth]{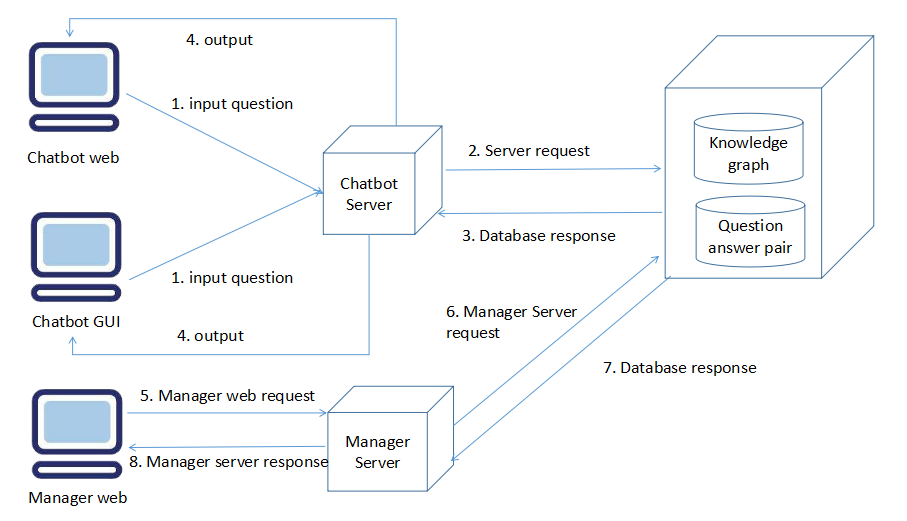}
\caption{The HHH System Architecture}
\label{fig1}
\end{figure*}

We review existing works that are related to the two problems above. 
\paragraph*{\bf Chatbots}
 Eliza was the first chatbot in the world developed at MIT Artificial Intelligence Laboratory by Joseph Weizenbaum in 1966 \cite{weizenbaum1966eliza}. However, Eliza cannot understand the question from the user. Parry was the first chatbot to pass the Turing Test created by psychiatrist Kenneth Colby in 1972 \cite{cerf1973parry}. Nevertheless, only 48\% of the psychiatrists can correctly figure out the real patient from the conversation.

 Ni et al. \cite{ni2017mandy,ni2018framework} tried to use the multiple-turn dialog decision tree to make a judgment for a patient. Helen et al. \cite{zhao2018finding} found that using transfer learning to transfer common scenarios from SQuAD to Bible QA can effectively improve the accuracy of the model on shorter context conversations. Dai, Z., etc. \cite{dai2016cfo} proposed a ``focused pruning method'' to reduce the candidate result space and make some improvements by using N-gram methods, which efficiently reduce the data noise. Wang, Y., etc. \cite{wang2018apva} proposed ``APVA'' to accurately predict the connection between the question entity and answer entity. Yih, S., etc. \cite{yih2015semantic} proposed a new semantic analysis framework when the question has been transferred and analyzed to query language, the new query will be related to the knowledge base. Yu, M., etc. \cite{yu2017improved} proposed a hierarchical RNN network by using residual learning to improve the performance in 2017. When there is an input question, it can detect the relation inside the knowledge base. Besides, they developed a simple KBQS system that integrates the entity linking and relation detector.

\paragraph*{\bf Knowledge Base Storage and Retrieval}
Cui et al. \cite{cui2017kbqa} built an open domain knowledge base question-and-answer system in 2017. They tried to design more templates from a billion scale QA corpora to better understand questions. However, they do not consider user intention with a knowledge graph so that the answer is limited by the template itself rather than capture the user intention. Lukovnikov, etc. \cite{lukovnikov2017neural} propose a model to capture useful information from different layers and combine the different characters of RNN and CNN. They have used RNN \cite{mikolov2010recurrent} to capture the semantic level connection and Attention \cite{vaswani2017attention} to follow the entity and relationship. However, RNN cannot capture the forward and backward context information.

\paragraph*{\bf Siamese based Semantic sentence similarity}
Mueller et al. have proposed a Siamese Long Short-Term Memory (LSTM) network to compute the semantic similarity between two variable-length sentences \cite{mueller2016siamese}. However, LSTM cannot detect the keywords from a sentence. Baziotis et al. \cite{baziotis2017datastories} proposed a Siamese architecture with Bidirectional Long Short-Term Memory (LSTM) networks with an attention mechanism. The model uses Bidirectional Long Short-Term Memory (LSTM) to capture both two-direction contexts. However, they consider the fully-connect (tanh) in the final layer to make the classification, which can cause over-fitting.

\section{System Architecture}\label{sec:system}
 We solve the two problems mentioned above by including a hybrid QA model in our chatbot framework, which combines a knowledge graph to manage a medical dataset and the HBAM to understand the text. The adoption of such a combined system is driven by the following motivations: 
 
 Firstly, a knowledge-based system holds some clear advantages in providing targeted responses to well-defined questions and thus is a convenient and reliable approach in implementing a question-answering system in knowledge-centric domains such as medical fields. A predominant type of medical question seeks explanations of specific symptoms that have rather specialized knowledge, and a knowledge-based system can quickly return the desired results upon requests. Furthermore, certain questions require a certain amount of logical reasoning, and these are, e.g., deriving the cause of certain illness, which can also be solved using a knowledge representation approach such as RDF queries, as RDF triples in a knowledge graph can well represent the complex connections between entities. Therefore, it is natural to adopt a knowledge graph as an integral part of a question-and-answering system. 
 

Secondly, a knowledge-based system can sometimes be too rigid in a conversational context. A patient may not be able to use the vast amount of domain-specific and accurate keywords in formulating a question, but rather, they resort to a casual and even layman's language. The knowledge graph contains a fixed set of knowledge, and when the system fails to match a question with an RDF triple, a knowledge-based system may fail to provide a meaningful answer. Thus, it is beneficial to go beyond merely encoding knowledge explicitly by RDF triples, but preferably using an alternative, data-driven approach. Given this limitation, we propose a neural-based model, namely, HBAM, which provides a more flexible model for various situations. The most frequently-used questions in the conversation model can be filtered first, followed by dialogue understanding. Furthermore, when the knowledge graph cannot be parsed and matched to the appropriate problem, the method of comparing the similarities is used to find the most similar problem in the question-and-answer the knowledge base. The utilization of HBAM is expected to improve the dialogue quality of our system.

%
 Figure~\ref {fig1} summarizes the system architecture of HHH: 
 \begin{itemize}
 \item two chatbot clients (website and GUI) connect with a chatbot server; 
 \item one manager client communicates with a manager server; 
 \item a hybrid QA model aims to respond to the messages from the chatbot server with two datasets (a knowledge graph, and a medical QA pair dataset) that are managed through the manager server. 
\end{itemize}

The knowledge graph is developed by Neo4j\footnote{https://neo4j.com/\label{b_neo4j}} with data from the Health Navigator New Zealand\footnote{https://www.healthnavigator.org.nz/apps-videos/b/\label{b1_1}}, common illnesses and symptom\footnote{https://www.nhsinform.scot/illnesses-and-conditions/a-to-z\label{b2_1}} and common diseases and conditions\footnote{https://www.medicinenet.com/diseases\_and\_conditions/article.htm\label{b3_1}}. The QA pair dataset\footnote{https://github.com/LasseRegin/medical-question-answer-data} is generated in 2017, originally from eHealth Forum\footnote{https://ehealthforum.com/health/health\_forums.html}, Question Doctors\footnote{https://questiondoctors.com/blog/} and WebMD\footnote{https://www.webmd.com/a-to-z-guides/qa} (HealthTap and iCliniq are not used). 
HBAM (which will be presented in Section~\ref{sec:HBAM}) will be used to find the best match questions from this QA pair dataset and return the answers to the user.

\begin{figure*}[]
\centering
\includegraphics[width=0.79\textwidth]{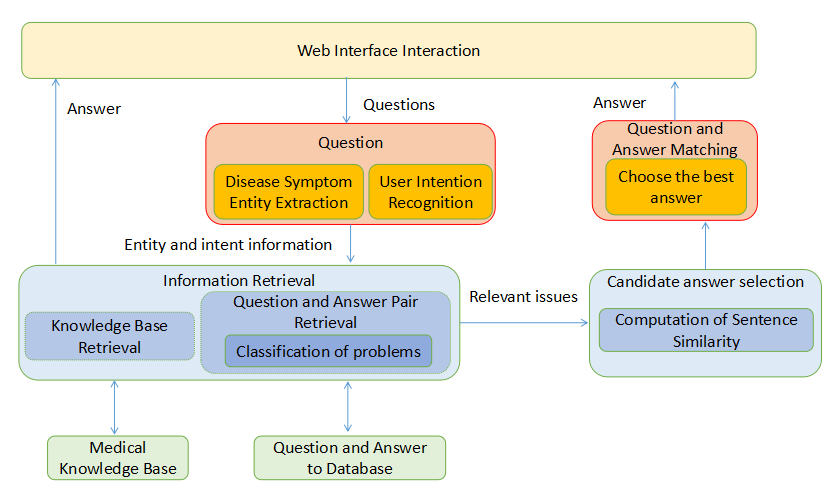}
\caption{A question-and-answer framework that combines knowledge graph and HBAM}
\label{fig5}
\end{figure*}

 Figure~\ref{fig5} shows the hybrid QA model in the Chatbot framework. When a user's question is given as input,  it can be processed by our two QA retrieval modules.  
 \begin{enumerate}
 \item The information from ``Web Interface Interaction'' will be transferred into the information retrieval module, which first tries to retrieve the answer from our two datasets. If the answer can be extracted directly from the knowledge graph dataset, the information retrieval module can retrieve and return the answer.
 \item If, on the other hand, the required answer cannot be found from the knowledge graph due to the limitation of the scale of the dataset. In this case, the question will be transferred into the question-answer pair retrieval module. Here we use HBAM to check the semantic similarity of the user's question and the questions from the question-answer pair dataset. The top $k$ most similar questions will be returned as the answer set.
\end{enumerate}
In the next two sections, we describe how we implement the two models for our medical chatbot system. 

\section{The Knowledge Graph Architecture} \label{sec:KG}
\subsection{Storage Scale of the Neo4j Graph Database}
 The system knowledge graph which contains \begin{itemize}
 \item 3 entities (department, disease, and symptom), 
 \item 6 properties (name, description, cause, prevent, accompany, cure\_way), and 
 \item 5 relationships (have\_symptom, accompany\_with, disease\_prevent, disease\_cause, disease\_cureway).
 \end{itemize}
 There are about 3,500 entities (which include 675 diseases and 2825 symptoms) and 4,500 relationships. The relationship includes the relationship between the diseases, symptoms, and the other 6 properties.
\subsection{The Process of Selecting Answers from the Graph Database}
The whole process can be divided into five steps: 1. User input question 2. Extract entity with D\&S Extractor 3. Get user intention by Intention Recognizer 4. Answer Selection 5. return the answer.

\begin{figure*}[t]
\centering
\includegraphics[width=0.8\textwidth]{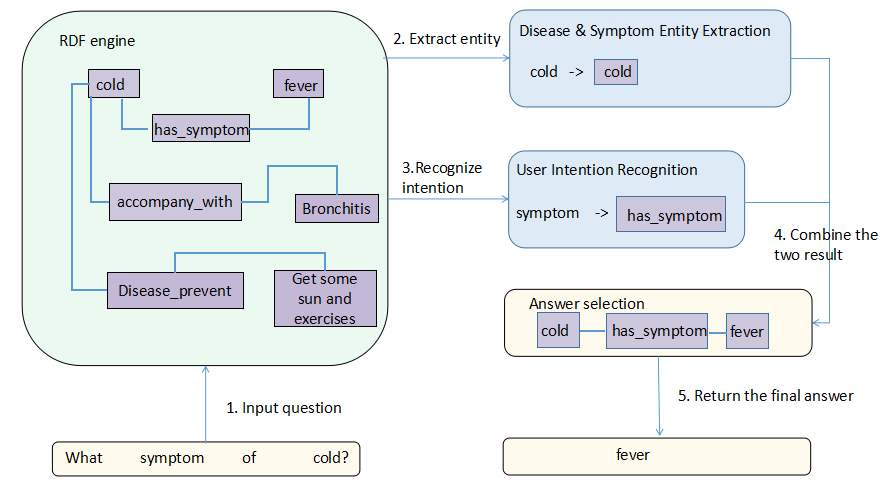}
\caption{Answer feedback from knowledge graph}
\label{fig11}
\end{figure*}

As an example, Figure~\ref{fig11} shows how the word ``cold'' is detected as a disease keyword by the Entity Extractor, how the intention ``has\_symptom'' is recognized by Intention Recognizer, and how the answer ``fever'' is selected.

\subsection{Design of Problem Analysis Module}
\begin{figure*}[b]
\centering
\includegraphics[width=0.88\textwidth]{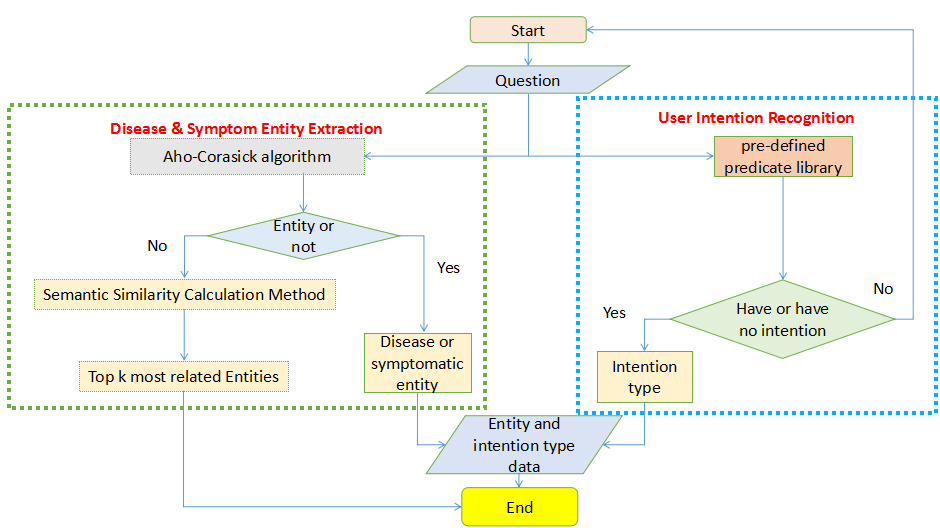}
\caption{Entity detection and Intention recognition}
\label{fig5_1}
\end{figure*}

Figure \ref{fig5_1} shows the disease symptom entity extraction functionality of the system. This function extracts the disease keywords from a medical keywords dictionary and is performed using the Aho-Corasick algorithm \cite{aho1975efficient}. If the Aho-Corasick algorithm does not identify the disease and symptom entities in the given question, it will enter the semantic similarity calculation module, and search for the most similar k entities in semantics. The user interaction recognition is to predict the user intention by some pre-defined predicate libraries. If the pre-defined predicate libraries do not recognize the intention of the given question, it will prompt the user to ask again, which means the system cannot understand the meaning of the question. Overall, Six typical questions can be answered by our system, based on the five relationships. 

\section{Hierarchical BiLSTM Attention Model} \label{sec:HBAM}
 The diagram of the new hierarchical BiLSTM Attention model we proposed is shown below in Figure~\ref{fig2}.

\begin{figure*}[t]
\centering
\includegraphics[width=1\textwidth]{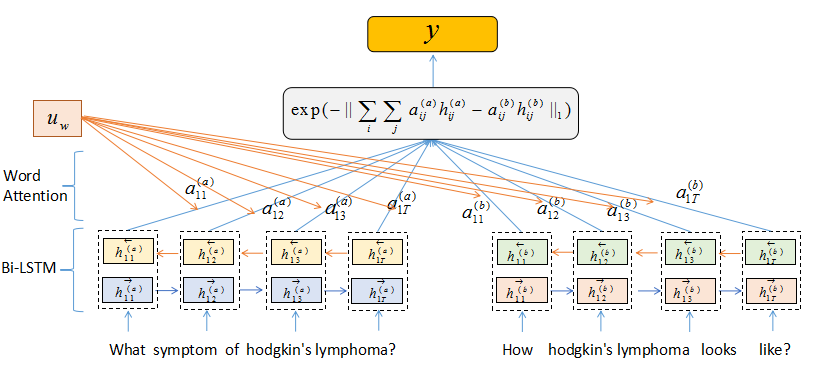}
\caption{Hierarchical BiLSTM Attention Model}
\label{fig2}
\end{figure*}
 It is designed for semantic similarity comparison. The whole structure based on a Siamese LSTM framework \cite{mueller2016siamese}. We apply one BiLSTM layer and one word attention layer into the Siamese framework. The bottom left, and the right sentences represent user input query and the question from the QA dataset. The two questions will be represented by using word embedding \cite{mikolov2013distributed} firstly and then using BiLSTM \cite{schuster1997bidirectional} to form the whole sentence embedding based on the context. After that, each BiLSTM encoder will be multiplied by a word attention value, which can be assumed as a weight to highlight the key-point in a sentence. Context vector will be combined with attention to understanding the sentence representation $u _ { w }$ \cite{yang2016hierarchical}. Finally, the similarity value will be computed by the weighted sum of each hidden state value $h_{i j}$ multiply its attention value. The details will be shown in the subsections below.

\noindent\textbf{LSTM-based sequence encoder}
Long Short Term Memory networks (LSTMs) are proposed by Hochreiter \& Schmidhuber  \cite{hochreiter1997long}. 
 One of the critical things of LSTMs is the cell state. The state of the cell can be regarded as a sort of conveyor belt. With a few linear interactions, it goes straightly down the entire chain. It is straightforward for information to flow along with it unchanged. 

 The basic principle of LSTM can be divided into three steps. The first step shows which information will be forgotten by the cell state. The ``forget gate layer'' makes this choice by combing the $h_{t-1}$ and $x_t$, which means the value of the hidden layer at time $t-1$ and the value of the input layer at time $t$. $W_g$ means the weight matrix between the hidden layer and output layer, and $b_g$ means the bias vector. We can get the $g_t$ through this formula, which decides to filter out that unimportant information.
\begin{equation}
g _ { t } = \sigma \left( W _ { g } \cdot \left[ h _ { t - 1 } , x _ { t } \right] + b _ { g } \right)
\end{equation}
 The second step is to choose which new information will be stored in the state of the cell. The input gate layer determines which values are updated. A tanh layer produces a vector that can be added to consider whether the new candidate values $\tilde { C } _ { t }$ should be updated in the state. Then, these two values will be merged to generate a state update.

\begin{equation}
j _ { t } = \sigma \left( W _ { j } \cdot \left[ h _ { t - 1 } , x _ { t } \right] + b _ { j } \right)
\end{equation}
\begin{equation}
\tilde { C } _ { t } = \tanh \left( W _ { C } \cdot \left[ h _ { t - 1 } , x _ { t } \right] + b _ { C } \right)
\end{equation}

 The time that decides whether the old cell state $C _ { t - 1 }$ will be updated by the new cell state $C _ { t }$ is depending on the previous steps. The old state multiplied by $g _ { t }$, and the forgetting things will be chosen to forget earlier. After that $j _ { t } * \tilde { C } _ { t }$ will be added. So, there will be a new candidate value, ranged by how much each state value has been chosen to update.

 Finally, the output will be decided. The output will be filtered by the cell state. Then a sigmoid layer will be operated that is chosen which components of the cell state will be output. The cell state will be put through tanh ranging from $-1$ to $1$ and multiplied it by the sigmoid gate output so that the output will be decided by the chosen components.

\begin{equation}
q _ { t } = \sigma \left( W _ { q } \left[ h _ { t - 1 } , x _ { t } \right] + b _ { q } \right)
\end{equation}
\begin{equation}
h _ { t } = q _ { t } \times \tanh \left( C _ { t } \right)
\end{equation}

\noindent\textbf{Word Attention} Given a sentence $w _ { i t }$, $t \in [ 0 , T ]$. Firstly, each word of the sentence will be embedded by using a embedding matrix $\mathrm { W } _ { e }$. 

\begin{equation}
x _ { i t } = W _ { e } w _ { i t } , t \in [ 1 , T ]
\end{equation}

 We use Bidirectional LSTM \cite{tan2016improved} to capture both forward and reverse direction information of each word. The bidirectional LSTM contains forward LSTM $\vec{f}$ and reverse LSTM $\overleftarrow{f}$. 

\begin{equation}
\vec { h } _ { i t } = \vec { \operatorname { LSTM } } \left( x _ { i t } \right) , t \in [ 1 , T ]
\end{equation}
\begin{equation}
\vec { h } _ { t i } = \vec { \operatorname { LSTM } } \left( x _ { t i } \right) , t \in [ 1 , T ]
\end{equation}

 In order to represent those keywords in a sentence. We try to use Attention. Firstly, we feed the $h _ { i t }$ into the tanh function to get $u _ { i t }$ as a hidden representation of $h _ { i t }$. Secondly, we calculate the importance of each word $u _ { i t }$ and get a normalized importance weight $\alpha _ { i t }$ by using a softmax function. Then, we calculate the sentence vector $\boldsymbol { s } _ { \dot { \imath } }$ as a weight sum of each word with its weight.

\begin{equation}
u _ { i t } = \tanh \left( W _ { w } h _ { i t } + b _ { w } \right)
\end{equation}
\begin{equation}
\alpha _ { i t } = \frac { \exp \left( u _ { i t } \right) } { \sum _ { t } \exp \left( u _ { i t } \right) }
\end{equation}
\begin{equation}
s _ { i } = \sum _ { t } \alpha _ { i t } h _ { i t }
\end{equation}

\noindent \textbf{Similarity function}
\begin{equation}
    f \left( s _ { i } ^ { ( a ) } , s _ { j } ^ { ( b ) } \right) = \exp \left( - \left\| \sum _ { i } \sum _ { j } a _ { i j } ^ { ( a ) } h _ { i j } ^ { ( a ) } - a _ { i j } ^ { ( b ) } h _ { i j } ^ { ( b ) } \right\| _ { 1 } \right) \in [ 0,1 ]
\end{equation}
The formula is based on Manhattan distance. From this formula, the representation from two sentences can be represented by $s_i^{(a)} = \sum _ { i } \sum _ { j } a _ { i j } ^ { ( a ) } h _ { i j } ^ { ( a ) }$ and $s_j^{(b)} = \sum _ { i } \sum _ { j } a _ { i j } ^ { ( b ) } h _ { i j } ^ { ( b ) }$. $a_{ij}^{(a)}$ and $a_{ij}^{(b)}$ mean the attention value in both direction. $h_{ij}^{(a)}$ and $h_{ij}^{(b)}$ mean the hidden state value in both direction.

\begin{figure*}[t]
\centering
\includegraphics[width=0.643\textwidth]{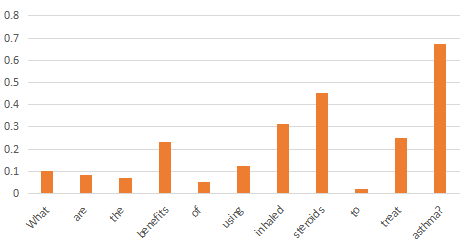}
\caption{The word attention distribution in the two sentences}
\label{fig7_2}
\end{figure*}

\begin{figure*}[t]
\centering
\includegraphics[width=0.84\textwidth]{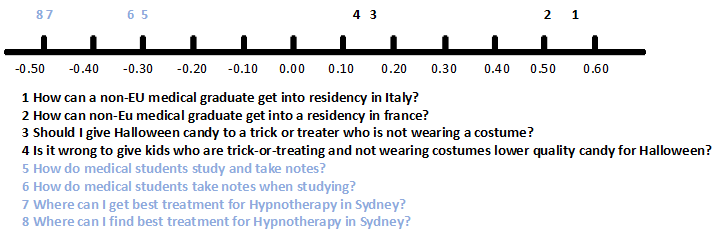}
\caption{The comparison of positive representation and the negative representation}
\label{fig7_1_1}
\end{figure*}
\begin{figure*}[b]
\centering
\includegraphics[width=0.7\textwidth]{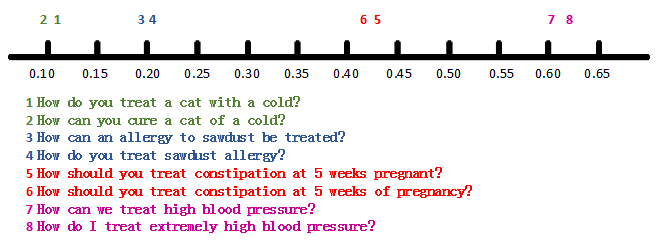}
\caption{The distribution of a different group of sentence representation}
\label{fig7_1_2}
\end{figure*}

\begin{figure*}[t]
\centering
\includegraphics[width=0.9\textwidth]{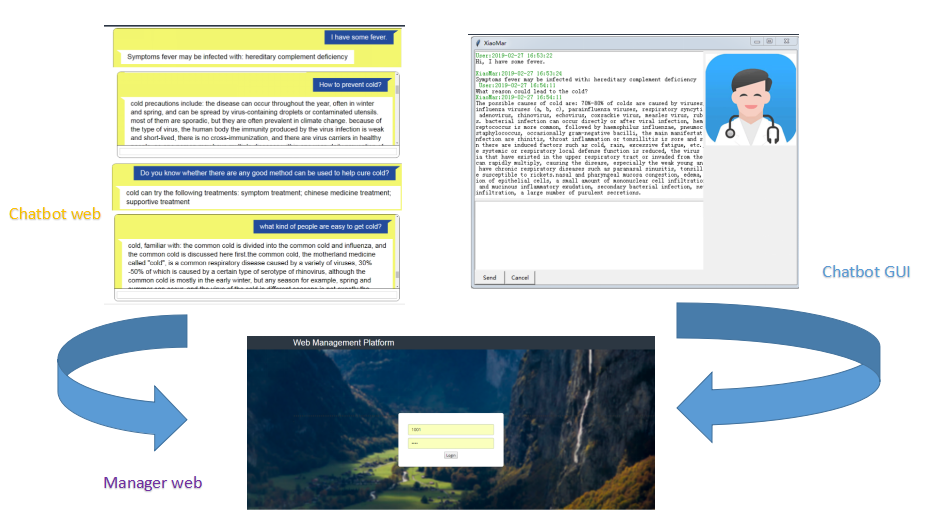}
\caption{Chatbot Website, GUI and Manager Website}
\label{fig7}
\end{figure*}

\section{Results and Analysis}\label{sec:results}

\subsection{Sample Results}
In order to explain the process of calculating the similarity of medical questions in HBAM model, we found that HBAM has an impressive text representation in understanding the word meaning and weight distribution of a sentence, as well as the distribution of the meaning of the sentence.

Each word in a sentence is represented by weight according to the word attention mechanism. As an example, Figure~\ref{fig7_2} shows a sentence ``\texttt{What are the benefits of using inhaled steroids to treat asthma?}'' The words ``\texttt{asthma}'' and ``\texttt{steroids}'' are given higher weight.

The sentence representation can be seen in Figure \ref{fig7_1_1} and Figure \ref{fig7_1_2}, respectively. The former shows the sentence vector distribution comparison between the positive representation and the negative representation.
The latter shows the distribution of a different group of sentence representation. The 1st and 2nd sentences both represent the meaning related to the cold. The 3rd and 4th sentences both represent the meaning related to the allergy. The 5th and 6th sentences both represent the meaning related to the pregnant, and the 7th and 8th sentences both represent the meaning related to the high blood pressure.

 Figure~\ref {fig7} shows a single-turn conversation example in the HHH. The image on the top-left displays the online chatbot interface, and the one on the top-right shows the chatbot GUI. They are both managed by the manager website (image at the bottom) - the Github link includes code and data \footnote{https://github.com/14H034160212/HHH-An-Online-Question-Answering-System-for-Medical-Questions}. In the following, we give further quantitative analysis on the system performance using two sets of experiments. 


\subsection{Experiment 1} \label{experiment}
\subsubsection{Train and Test Dataset}
The HBAM is trained with the data from Quora duplicate questions dataset \footnote{https://data.quora.com/First-Quora-Dataset-Release-Question-Pairs\label{b3_1_0}}. To filter out the medical subset from the dataset, we create a disease and symptom dictionary which contains medical keywords such as cold, obesity, weight loss, and low temperature according to two New Zealand medical website \footnote{https://www.nhsinform.scot/symptoms-and-self-help/a-to-z\label{b4_1_0}} \footnote{https://www.healthnavigator.org.nz/health-a-z/\label{b4_1_1}}. The number of disease and symptom keywords in the dictionary is 668 and 2367, respectively. With the dictionary, we collect nearly 70,000 medical-related records from the Quora dataset. For training the models faster and easier, we randomly select 10,000 records (positive: negative = 1:1) as the experiment data. The results of Experiment 2 in Section~\ref{supplementary} will demonstrate that the performances of the models for the remaining records are similar. 

The Quora duplicate questions dataset is an open domain sentence pair dataset. It has more than 400,000 tagged sentence pairs formatted like ``\texttt{text1 text2 is\_duplicate}'' means whether the two sentences are semantically similar. If they are semantically equal, the tag will be ``1'', otherwise ``0''. Some examples are list in table \ref{table6_3}.

\begin{table*}[]
\centering
\caption{Some examples in Quora medical subset}
\label{table6_3}
\begin{tabular}{|c|c|c|c|c|c|}
\hline
id     & qid1   & qid2   & question1 & question2 & is\_duplicate \\ \hline
130859 & 209926 & 209927 & \begin{tabular}[c]{@{}c@{}}How do you treat \\ a cat with a {\textcolor{red}{cold}}?\end{tabular} & \begin{tabular}[c]{@{}c@{}}How can you cure \\ a cat of a {\textcolor{red}{cold}}?\end{tabular} & 1             \\ \hline
82425  & 139763 & 133638 & \begin{tabular}[c]{@{}c@{}}How much medical evidence \\ is there in support of \\ the claim {\textcolor{cyan}{weed}} causes {\textcolor{cyan}{cancer}}?\end{tabular}                           & \begin{tabular}[c]{@{}c@{}}Does {\textcolor{cyan}{weed}} give \\ you {\textcolor{cyan}{lung cancer}}?\end{tabular}                         & 1             \\ \hline
261370 & 377490 & 377491 & \begin{tabular}[c]{@{}c@{}}How can an {\textcolor{olive}{allergy}} \\ to {\textcolor{olive}{sawdust}} be treated?\end{tabular} & \begin{tabular}[c]{@{}c@{}}How do you treat \\ {\textcolor{olive}{sawdust}} {\textcolor{olive}{allergy}}?\end{tabular}                       & 1             \\ \hline
...    & ...    & ...  & ...   & ... & ...           \\ \hline
\end{tabular}
\end{table*}

\begin{table*}[t]
\centering
\caption{Methods comparison}
\label{table6}
\begin{tabular}{|c|c|c|}
\hline
Methods & Average Evaluation Accuracy & \begin{tabular}[c]{@{}c@{}}Range of change by \\ 30 times experiments\end{tabular} \\ \hline
BERT \cite{devlin2018bert} & 78.2\% & (-1.8\%,+1.3\%) \\ \hline
MaLSTM \cite{hochreiter1997long} & 78.4\% & (-2.9\%,+2.0\%) \\ \hline
HBAM & {\bf 81.2\%} & (-2.4\%,+2.2\%) \\ \hline
\end{tabular}
\end{table*}

\begin{table*}[b]
\centering
\caption{Evaluation result for the three medical websites}
\label{table6_3_11}
\begin{tabular}{|c|c|c|c|}
\hline
Medical website name & Method name & Average predict accuracy & \begin{tabular}[c]{@{}c@{}}Range of change by \\ 10 times experiments\end{tabular} \\ \hline
\multirow{2}{*}{ehealthforumQAs} & BERT & 78.5\% & (-1.8\%,+1.1\%) \\ 
\cline{2-4}&   HBAM & {\bf 81.3\%} & (-1.2\%,+1.1\%) \\ 
\cline{2-4}&   MaLSTM & 78.4\% & (-2.9\%,+1.5\%) \\ \hline
\multirow{2}{*}{questionDoctorQAs} & BERT & 78.2\% & (-1.4\%,+0.9\%)\\ \cline{2-4} 
\cline{2-4}&   HBAM & {\bf 80.9\%}  & (-2.1\%,+2.5\%) \\ \cline{2-4} 
\cline{2-4}&   MaLSTM & 78.1\% & (-1.7\%,+1.9\%) \\ \hline
\multirow{2}{*}{webmdQAs} & BERT & 78.1\% & (-1.6\%,+0.9\%) \\ \cline{2-4} 
\cline{2-4}&   HBAM & {\bf 81.2\%} & (-1.2\%,+1.3\%) \\ \cline{2-4} 
\cline{2-4}&   MaLSTM & 78.5\% & (-1.5\%,+1.9\%) \\ \hline
\end{tabular}
\end{table*}

\subsubsection{Environment}
 We have experimented the deep learning models on Google Colab\footnote{https://colab.research.google.com/notebooks/welcome.ipynb\label{b18}} (Tesla K80 GPU, 12 GB RAM) to validate the semantic similarity between two sentences. The hyperparameters of HBAM includes the batch\_size is 1024, the n\_epoch is 9, the n\_hidden is 100, the embedding\_dim is 300 and the max\_seq\_length is 10, GoogleNews-vectors-negative300.bin.gz from Word2Vec\footnote{https://code.google.com/archive/p/word2vec/\label{b20}}, the activation function is tanh.

\subsubsection{Comparison}
To evaluate the performance of our system,  we compare it with two state-of-the-art sentence pair similarity algorithms, namely BERT and MALSTM \cite{mueller2016siamese}. BERT was proposed by Google in 2018 and has refreshed records in 11 NLP tasks, including Q\&A (SQuAD v1.1), reasoning (MNLI), and more. MALSTM was proposed by the MIT team in 2016 and has achieved excellent results in calculating the similarity of sentences. It is better than a few well-known sentence similarity comparison algorithms include Dependency Tree-LSTM, ConvNet, and more. Superior performance over these two benchmark means that our system would have achieved a level that is higher or on par with the current state-of-the-art methods. 

In Table~\ref {table6}, we display the results of mapping the medical-related words to query 10000 lines medical-related question pairs. We divide the dataset by 6:2:2 for training, validation, and testing in the BERT baseline model for fine-tuning. In other models, we use 9:1 for training and testing. It can be clearly seen that our HBAM has the best performance to check the duplication of two text sentences.

\subsection{Experiment 2} \label{supplementary}
We also perform a second experiment on the remaining more than 50,000 medical sentence pairs as well. We separately select thousands of tags from the three kinds of datasets: ehealthforumQAs, questionDoctorQAs, and webmdQAs, respectively. The tags of each dataset are extracted as the keywords to take the intersection with the disease symptom keyword dictionary. Then the intersection results are used to search for matching sentence pairs in the remaining 50,000 medical sentence pairs. Finally, the first 1000 matched sentence pairs are taken out for each dataset, and 10 times evaluation results are obtained, respectively.

Table \ref{table6_3_11} shows the evaluation results by experiment 10 times. From the three tables, we believe that the HBAM performs better prediction performance in the three test cases.

In Experiment 2, we reuse the models trained from Section~\ref{experiment}, but test with different datasets. Nevertheless, it has turned out that the accuracy of the three models has not changed in a significant way.


\section{Conclusion and Future Work}\label{sec:conclusion}
 In this paper, we propose a chatbot framework based on a knowledge graph and a text representation and similarity model. The advantage of the knowledge graph lies in that it utilizes structured storage so that it may help easy maintenance and retrieval of domain-specific knowledge. While the advantage of the attention model utilizes deep learning to represent better and comprehend natural language questions. Therefore, we develop a system that combines the advantages of both models by integrating a knowledge graph with a neural-based model. We compare the ability to achieve the text-similarity between some state-of-the-art NLP models and our new HBAM. We consider the scenarios of single-turn question-and-answer dialogue, use the method of the knowledge graph, and combine deep learning methods to present data well. We speculate that one reason that HBAM is better than MaLSTM is by adding the attention layer to help capture the medical keywords from a sentence. Besides, two possible reasons for HBAM's superior performance over BERT is because of BERT is pre-trained based on a general word embedding and the 12-layer transformer which could cause overfitting when we try to capture and understand the medical keywords from a sentence.


As future work, we foresee the potential of chatbot technologies to play a much more significant role in the medical domain. For example, a software chatbot can be deployed in the real-world to become home healthcare robots or hospital medical inquiry robots. 
From the application point of view, the paper only considered the single-turn question-and-answer mechanism. An important future direction is to add user profiles into the system and provide a more precise medical assistant to each specific user. Besides, we can combine the data mining method and predict the potential diseases in a region of the population. We plan to recruit some participants to help to evaluate our medical QA system. Also, we hope in the future our chatbot framework can have a chance to be applied in other domains besides healthcare.

\section{Acknowledgement}\label{sec:acknowledgement}
The authors would like to thank Yang Chen who provides many useful suggestions. This research was supported by summer scholarship funding from the Precision Driven Health research partnership.


\balance
\bibliographystyle{ACM-Reference-Format}
\bibliography{sample-base}


\end{document}